\documentclass{Interspeech}
\usepackage{multirow}
\usepackage{graphicx}
\usepackage{caption}

\interspeechcameraready

\title{ViCocktail: Automated Multi-Modal Data Collection for \\Vietnamese Audio-Visual Speech Recognition}

\author[affiliation={1*}]{Thai-Binh}{Nguyen}
\author[affiliation={2*}]{Thi Van}{Nguyen}
\author[affiliation={2*}]{Quoc Truong}{Do}
\author[affiliation={3}]{Chi Mai}{Luong}
\affiliation{VietAI Research}{New Turing Institute}{Vietnam}
\affiliation{}{Phenikaa University}{Vietnam}
\affiliation{}{Institute of Information Technology}{Vietnam}
\email{nguyenvulebinh@gmail.com}


\usepackage{comment}

\begin{document}

\maketitle

\begingroup 
\makeatletter 
\renewcommand{\thefootnote}{\fnsymbol{footnote}} 
\setcounter{footnote}{0} 

\stepcounter{footnote} 
\footnotetext{Corresponding author.}
\makeatother
\endgroup

\begin{abstract}

Audio-Visual Speech Recognition (AVSR) has gained significant attention recently due to its robustness against noise, which often challenges conventional speech recognition systems that rely solely on audio features. Despite this advantage, AVSR models remain limited by the scarcity of extensive datasets, especially for most languages beyond English. Automated data collection offers a promising solution. This work presents a practical approach to generate AVSR datasets from raw video, refining existing techniques for improved efficiency and accessibility. We demonstrate its broad applicability by developing a baseline AVSR model for Vietnamese. Experiments show the automatically collected dataset enables a strong baseline, achieving competitive performance with robust ASR in clean conditions and significantly outperforming them in noisy environments like cocktail parties. This efficient method provides a pathway to expand AVSR to more languages, particularly under-resourced ones.
    
\end{abstract}

\section{Introduction}

The task of audio-visual speech recognition (AVSR) is not new, with its roots tracing back to the discovery of the McGurk effect \cite{mcgurk1976hearing} in 1976, which demonstrated the strong interaction between auditory and visual speech perception. Inspired by this interplay, AVSR systems \cite{41402, duchnowski94_icslp, waibel1996multimodal} have been developed to leverage visual cues for enhancing speech perception, particularly in noisy environments. Since then, numerous AVSR datasets have been developed to support research in this field. While English datasets such as LRS2 \cite{8099850}, LRS3-TED \cite{afouras2018lrs3tedlargescaledatasetvisual}, and LSVSR \cite{shillingford19_interspeech} remain predominant, a growing number of datasets have been introduced for other languages, including Arabic (AVAS \cite{6920467}, AVSD \cite{ELREFAEI2019400}), Chinese (CN-CVS \cite{10095796}, NSTDB \cite{Chen2020LipreadingWD}, CMLR \cite{10.1145/3338533.3366579}), Russian (RUSAVIC \cite{ivanko-etal-2022-rusavic}, HAVRUS \cite{Verkhodanova2016HAVRUSCH}), Spanish (VLRF \cite{7961743}), and Turkish (VLRDT \cite{data8010015}).

There are two main approaches to developing an AVSR dataset. The first involves recording in controlled laboratory conditions to address specific AVSR challenges, such as vocabulary components or noise robustness. Examples include AVAS, AVSD, RUSAVIC, HAVRUS, and VLRF. These datasets are typically clean, manually labeled, but small in size. The second approach collects videos from uncontrolled sources like TV broadcasts or YouTube, as seen in LRS2, LRS3-TED, LSVSR, CN-CVS, NSTDB, CMLR, RUSAVIC, and VLRDT. These datasets are often larger but less clean and typically rely on multi-stage pipelines, like that proposed in \cite{10.1007/978-3-319-54184-6_6}, for automated collection and pre-processing.

Despite the growing number of AVSR datasets, the availability of resources for model development remains limited. For English, the largest dataset, LSVSR, contains 3,886 hours of speech but is not publicly accessible. LRS3-TED, with 475 hours, is also unavailable for download at the time of this study, leaving only LRS2 with 224 hours as a usable resource. For other languages, the available hours are even fewer, and most datasets are not publicly accessible. This scarcity of open datasets significantly hinders the development and benchmarking of AVSR models. While the multi-stage pipeline has been used in many studies, a fully comprehensive and open-source solution is still lacking.

Recent advancements in AVSR have been driven by deep learning, particularly end-to-end models like Transformer \cite{shi22_interspeech, rouditchenko24_interspeech} and Conformer \cite{10096889}. These data-hungry approaches benefit immensely from large datasets, enabling techniques like self-supervised pre-training (AV-HuBERT \cite{shi22_interspeech}) or leveraging pre-trained Automatic Speech Recognition (ASR) models to generate transcriptions for vast unlabeled AV datasets \cite{10096889}. In English, the availability of large-scale datasets allows for a range of approaches, from self-supervised learning to training from scratch. However, for other languages with smaller datasets, such as CN-CVS, CMLR, and RUSAVIC, studies often rely on training from scratch. In ASR, many studies have explored adapting pre-trained models from high-resource languages to low-resource ones, such as XLS-R \cite{babu22_interspeech, xu22b_interspeech}. We hypothesize that AVSR can similarly benefit from this approach, as visual features may be more language-agnostic and transferable across languages.

In this study, we develop a fully automated data processing pipeline, inspired by \cite{10.1007/978-3-319-54184-6_6}, to create an AVSR dataset from raw video. The pipeline extracts lip regions of interest, audio, and transcriptions and is designed to scale across different languages. We then explore adapting pre-trained AVSR models from high-resource languages, such as English or multilingual models, to a low-resource language, testing the hypothesis that AVSR models can efficiently transfer knowledge between languages. Our experiments focus on Vietnamese, as this is the first AVSR system developed for the language. Vietnamese benefits from a strong ASR system for automated transcription and has unique phonetic and linguistic characteristics, including six tones. Visual cues, such as lip movements and facial expressions, can help distinguish tonal differences and improve recognition accuracy. To ensure reliable evaluation, we manually annotate the data instead of relying solely on automatic transcription.

Our contributions are as follows: (1) We open-source a fully automated data processing pipeline for creating AVSR datasets. (2) We develop and open-source the first-ever AVSR dataset for Vietnamese, including a high-quality test set for rigorous evaluation. (3) We explore knowledge transfer from pre-trained models to establish a strong baseline for Vietnamese AVSR. Our experiments show that with this adaptation, just 269 hours of AVSR data can match a strong ASR baseline on clean audio and significantly outperform it in noisy conditions.




\section{Methodology}
\subsection{ViCocktail dataset}

In this section, we describe the multi-stage pipeline for automatically generating a dataset for AVSR model. Through this pipeline, we have collected about 269 hours of video of spoken sentences and phrases along with the corresponding facetrack from an initial dataset of about 528.8 hours of video. We use a variety of Youtube channels such as talk show, vlogs. The selection of channels is deliberately based on three main reasons: (1) a diverse range of speakers, (2) close-up shots of the speaker's face, and (3) infrequent shot changes. Therefore there are more full sentences with continuous facetracks.

\begin{figure}[ht]
\centering
\includegraphics[width=0.47\textwidth]{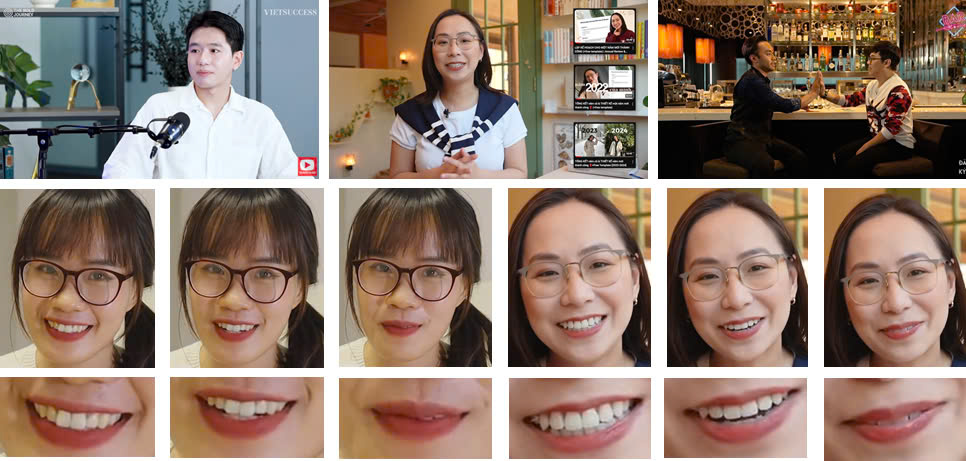}
\caption{\textbf{Top}: Original still images from videos used in the making of the dataset. \textbf{Bottom}: The mouth motions for ‘chào’ from two different speakers.}
\label{fig:yourlabel}
\vspace{-1em}
\end{figure}

The processing pipeline is summarised in Figure \ref{pipeline}:

\begin{figure}[ht]
  \centering
  \includegraphics[width=0.47\textwidth, trim=0.5cm 0.5cm 0.5cm 0.5cm, clip]{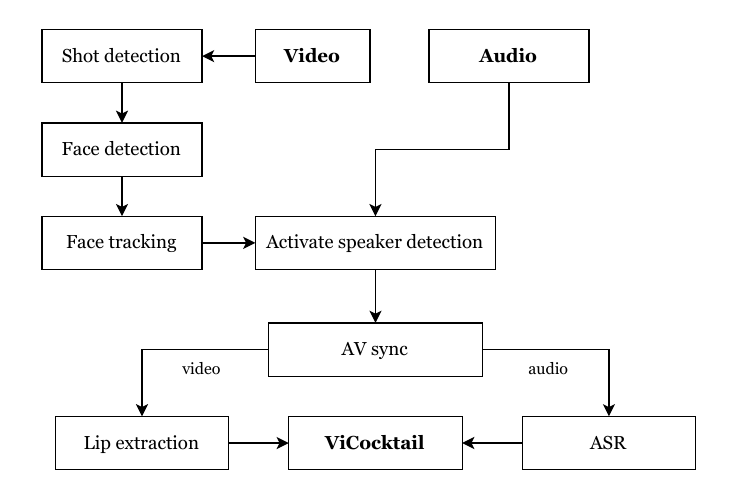}
  \caption{Pipeline to generate the ViCocktail dataset.}
  \vspace{-0.5em}
  \label{pipeline}
\end{figure}

\noindent \textbf{Video preparation.} A CNN face detector based on the Single Shot Scale-invariant Face Detector (S3FD) \cite{zhang2017s3fd} is used to detect face appearances in the individual frames. Then, shot boundaries are identified using a content-based scene detection algorithm that analyzes frame differences. Within each shot, face tracking is performed using the Intersection Over Union (IOU) algorithm to ensure temporal consistency. 

\noindent \textbf{Active Speaker Detection and AV sync.} In videos, the audio and the video streams can be out of sync, which can cause problems when the facetrack corresponding to a sentence is being extracted. We use two models: Active Speaker Detection \cite{Liao_2023_CVPR} (ASD) and SyncNet \cite{Chung2016OutOT} to address this problem. The ASD model relies on both audio and visual features to determine when a person in the video is actually speaking. Then, the SyncNet model analyzes the delay between lip movements and the corresponding audio to check if the sound is coming from the person in the video or from an external source. SyncNet provides information about the delay and the confidence level of the audio-visual sync, allowing adjustments to improve video quality and remove unsuitable videos.

\noindent \textbf{Lip extraction.} The mouth region is extracted from video frames using Dlib’s\cite{king2009dlib} face detection and landmark prediction models. First, the face is detected using a combination of a frontal face detector and a CNN-based detector. Then, 68 facial landmarks are predicted, focusing on the mouth area (landmarks 48-67). The extracted mouth region is then resized and normalized. Finally, the processed mouth-region video is saved and synchronized with the corresponding audio using FFmpeg.

\noindent \textbf{Audio and text preparation.} Since subtitles are unavailable, we use a Vietnamese ASR model to generate text labels. We experimented with various open-source and proprietary ASR systems to identify the most effective one for automatic labeling. Among them, a variant of the Wav2Vec2 model proposed by \cite{Thai_Binh_Nguyen_wav2vec2_vi_2021}, pre-trained on 13,000 hours of unlabeled data and fine-tuned on a 3,000-hour ASR dataset following the approach in \cite{10446589, 10889116}, demonstrated the best performance. We selected this model for automatically generating text labels for our ViCocktail dataset. 

\noindent \textbf{Datasets for testing model.} In addition to the data collection criteria, the test dataset is selected from YouTube videos to ensure that the speakers do not overlap with the training set. It also includes videos with background noise, such as multiple people speaking at the same time or environmental noise. To ensure an accurate evaluation of the model, all text labels in the test dataset are manually annotated.
\vspace{-0.5em}

\begin{table}[ht]
\centering
\caption{Overview of the ViCocktail dataset}
\vspace{-0.75em}
\label{tab:dataset-fullborder}
\begin{tabular}{|l|c|r|r|r|}
\hline
\textbf{Set} & \textbf{Dates} & \textbf{\# Utter.} & \textbf{Speaker} & \textbf{Duration} \\ \hline
Train & 12/24 - 01/25 & 194,073  & 2375 & 269h \\ \hline
Test  & 01/25  & 1,167 & 36 & 2h \\ \hline
\textbf{All} & & 195,240  & 2411 & 271h \\ \hline
\end{tabular}
\vspace{-1.5em}
\end{table}

\begin{figure*}[ht]
  \centering
  \includegraphics[width=1.0\linewidth]{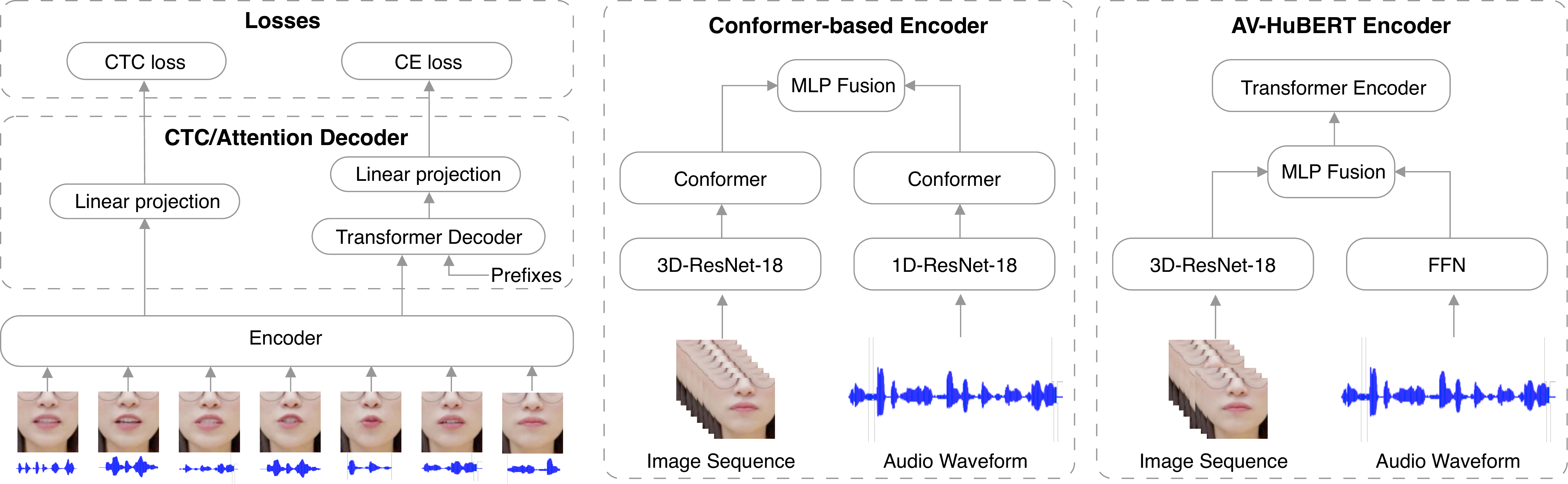}
  \caption{\textbf{Left:} Overview of the AVSR model, which consists of an encoder and a CTC/Attention decoder. Training utilizes two loss functions: CTC loss and cross-entropy loss. \textbf{Middle and Right:} Two variants of the encoder.}
  \vspace{-1.5em}
  \label{fig:avsr_model_overview}
\end{figure*}

\begin{table*}[h]
\caption{WER (\%) of models on the ViCocktail test set.}
\label{tab:experiment}
\centering
\begin{tabular}{cccccccccccc}
\hline
\multirow{2}{*}{Model}                                                             & \multirow{2}{*}{\begin{tabular}[c]{@{}c@{}}Fine\\ tuning\end{tabular}} & \multirow{2}{*}{\begin{tabular}[c]{@{}c@{}}Model \\ ID\end{tabular}} & \multirow{2}{*}{\begin{tabular}[c]{@{}c@{}}Pre-trained\\ language\end{tabular}} & \multirow{2}{*}{Modality}                                               & \multirow{2}{*}{\begin{tabular}[c]{@{}c@{}}Interferer \\ Speakers\end{tabular}} & \multicolumn{5}{c}{SNR (dB)}                                                                                                              & \multirow{2}{*}{Avg}                       \\
                                                                                   &                                                                        &                                                                      &                                                                                 &                                                                         &                                                                                 & -5                        & 0                         & 5                         & 10                        & $\infty$                       &                                            \\ \hline
\multirow{9}{*}{\begin{tabular}[c]{@{}c@{}}AV-HuBERT\\ CTC/Attention\end{tabular}} & \multirow{9}{*}{x}                                                     & \multirow{3}{*}{AV1}                                                 & \multirow{3}{*}{English}                                                        & \multirow{9}{*}{\begin{tabular}[c]{@{}c@{}}Audio\\ Visual\end{tabular}} & 0                                                                               &                           &                           &                           &                           & 9.40                      & \multirow{3}{*}{\textbf{12.78}}            \\
                                                                                   &                                                                        &                                                                      &                                                                                 &                                                                         & 1                                                                               & \textbf{15.34}            & \textbf{11.82}            & \textbf{12.12}            & \textbf{10.97}            &                           &                                            \\
                                                                                   &                                                                        &                                                                      &                                                                                 &                                                                         & 2                                                                               & \textbf{18.39}            & \textbf{12.87}            & \textbf{13.19}            & \textbf{10.96}            &                           &                                            \\ \cline{7-12} 
                                                                                   &                                                                        & \multirow{3}{*}{AV2}                                                 & \multirow{3}{*}{Multilingual}                                                   &                                                                         & 0                                                                               &                           &                           &                           &                           & 14.28                     & \multirow{3}{*}{20.22}                     \\
                                                                                   &                                                                        &                                                                      &                                                                                 &                                                                         & 1                                                                               & 24.92                     & 18.62                     & 18.70                     & 16.89                     &                           &                                            \\
                                                                                   &                                                                        &                                                                      &                                                                                 &                                                                         & 2                                                                               & 29.62                     & 20.35                     & 20.79                     & 17.80                     &                           &                                            \\ \cline{7-12} 
                                                                                   &                                                                        & \multirow{3}{*}{AV3}                                                 & \multirow{3}{*}{From scratch}                                                   &                                                                         & 0                                                                               &                           &                           &                           &                           & 18.60                     & \multirow{3}{*}{26.98}                     \\
                                                                                   &                                                                        &                                                                      &                                                                                 &                                                                         & 1                                                                               & 33.27                     & 24.61                     & 25.15                     & 22.29                     &                           &                                            \\
                                                                                   &                                                                        &                                                                      &                                                                                 &                                                                         & 2                                                                               & 41.12                     & 27.11                     & 27.83                     & 22.82                     &                           &                                            \\ \hline
\multirow{3}{*}{\begin{tabular}[c]{@{}c@{}}Conformer\\ CTC/Attention\end{tabular}} & \multirow{3}{*}{x}                                                     & \multirow{3}{*}{AV4}                                                 & \multirow{3}{*}{English}                                                        & \multirow{3}{*}{\begin{tabular}[c]{@{}c@{}}Audio\\ Visual\end{tabular}} & 0                                                                               &                           &                           &                           &                           & 14.4                      & \multirow{3}{*}{28.22}                     \\
                                                                                   &                                                                        &                                                                      &                                                                                 &                                                                         & 1                                                                               & 30.36                     & 29.89                     & 26.24                     & 25.34                     &                           &                                            \\
                                                                                   &                                                                        &                                                                      &                                                                                 &                                                                         & 2                                                                               & 37.00                     & 33.81                     & 28.38                     & 28.56                     &                           &                                            \\ \hline
\multirow{3}{*}{Wav2Vec2 \cite{Thai_Binh_Nguyen_wav2vec2_vi_2021}}                                                          & \multirow{3}{*}{}                                                      & \multirow{3}{*}{A1}                                                  & \multirow{3}{*}{Vietnamese}                                                     & \multirow{3}{*}{Audio}                                                  & 0                                                                               &                           &                           &                           &                           & \textbf{7.53}             & \multirow{3}{*}{28.38}                     \\
                                                                                   &                                                                        &                                                                      &                                                                                 &                                                                         & 1                                                                               & 55.62                     & 24.48                     & 21.07                     & 12.38                     &                           &                                            \\
                                                                                   &                                                                        &                                                                      &                                                                                 &                                                                         & 2                                                                               & 70.01                     & 27.41                     & 24.16                     & 12.79                     &                           &                                            \\ \hline
\multirow{3}{*}{PhoWhisper \cite{le2024phowhisper}}                                                        & \multicolumn{1}{l}{\multirow{3}{*}{}}                                  & \multirow{3}{*}{A2}                                                  & \multicolumn{1}{l}{\multirow{3}{*}{Vietnamese}}                                 & \multirow{3}{*}{Audio}                                                  & 0                                                                               & \multicolumn{1}{l}{}      & \multicolumn{1}{l}{}      & \multicolumn{1}{l}{}      & \multicolumn{1}{l}{}      & \multicolumn{1}{l}{13.44} & \multicolumn{1}{l}{\multirow{3}{*}{34.73}} \\
                                                                                   & \multicolumn{1}{l}{}                                                   &                                                                      & \multicolumn{1}{l}{}                                                            &                                                                         & 1                                                                               & \multicolumn{1}{l}{64.74} & \multicolumn{1}{l}{30.84} & \multicolumn{1}{l}{26.82} & \multicolumn{1}{l}{17.47} & \multicolumn{1}{l}{}      & \multicolumn{1}{l}{}                       \\
                                                                                   & \multicolumn{1}{l}{}                                                   &                                                                      & \multicolumn{1}{l}{}                                                            &                                                                         & 2                                                                               & \multicolumn{1}{l}{78.00} & \multicolumn{1}{l}{33.84} & \multicolumn{1}{l}{28.95} & \multicolumn{1}{l}{18.51} & \multicolumn{1}{l}{}      & \multicolumn{1}{l}{}                       \\ \hline
\multirow{9}{*}{\begin{tabular}[c]{@{}c@{}}AV-HuBERT\\ CTC/Attention\end{tabular}} & \multirow{9}{*}{x}                                                     & \multirow{3}{*}{A3}                                                  & \multirow{3}{*}{English}                                                        & \multirow{9}{*}{Audio}                                                  & 0                                                                               &                           &                           &                           &                           & 12.55                     & \multirow{3}{*}{38.78}                     \\
                                                                                   &                                                                        &                                                                      &                                                                                 &                                                                         & 1                                                                               & 39.65                     & 32.46                     & 42.26                     & 38.78                     &                           &                                            \\
                                                                                   &                                                                        &                                                                      &                                                                                 &                                                                         & 2                                                                               & 57.67                     & 38.85                     & 48.55                     & 38.25                     &                           &                                            \\ \cline{7-12} 
                                                                                   &                                                                        & \multirow{3}{*}{A4}                                                  & \multirow{3}{*}{Multilingual}                                                   &                                                                         & 0                                                                               &                           &                           &                           &                           & 14.43                     & \multirow{3}{*}{40.72}                     \\
                                                                                   &                                                                        &                                                                      &                                                                                 &                                                                         & 1                                                                               & 40.67                     & 35.53                     & 43.73                     & 41.01                     &                           &                                            \\
                                                                                   &                                                                        &                                                                      &                                                                                 &                                                                         & 2                                                                               & 57.98                     & 41.87                     & 49.67                     & 41.60                     &                           &                                            \\ \cline{7-12} 
                                                                                   &                                                                        & \multirow{3}{*}{A5}                                                  & \multirow{3}{*}{From scratch}                                                   &                                                                         & 0                                                                               &                           &                           &                           &                           & 47.64                     & \multirow{3}{*}{63.89}                     \\
                                                                                   &                                                                        &                                                                      &                                                                                 &                                                                         & 1                                                                               & 69.83                     & 59.67                     & 62.83                     & 58.73                     &                           &                                            \\
                                                                                   &                                                                        &                                                                      &                                                                                 &                                                                         & 2                                                                               & 82.40                     & 65.37                     & 67.40                     & 61.12                     &                           &                                            \\ \hline
\multirow{3}{*}{\begin{tabular}[c]{@{}c@{}}AV-HuBERT\\ CTC/Attention\end{tabular}} & \multirow{3}{*}{x}                                                     & V1                                                                   & English                                                                         & \multirow{3}{*}{Visual}                                                 &                                                                                 & \multicolumn{6}{c}{41.34}                                                                                                                                                              \\
                                                                                   &                                                                        & V2                                                                   & Multilingual                                                                    &                                                                         &                                                                                 & \multicolumn{6}{c}{46.96}                                                                                                                                                              \\
                                                                                   &                                                                        & V3                                                                   & From scratch                                                                    &                                                                         &                                                                                 & \multicolumn{6}{c}{53.00}                                                                                                                                                              \\ \hline
\multirow{2}{*}{\begin{tabular}[c]{@{}c@{}}Conformer\\ CTC/Attention\end{tabular}} & \multirow{2}{*}{x}                                                     & V4                                                                   & English                                                                         & \multirow{2}{*}{Visual}                                                 &                                                                                 & \multicolumn{6}{c}{42.16}                                                                                                                                                              \\
                                                                                   &                                                                        & V5                                                                   & \multicolumn{1}{l}{From scratch}                                                &                                                                         &                                                                                 & \multicolumn{6}{c}{64.51}                                                                                                                                                              \\ \hline
\end{tabular}
\vspace{-1.5em}
\end{table*}

\subsection{Models}

We adopt two off-the-shelf architectures, as shown in Figure \ref{fig:avsr_model_overview}. Both follow a sequence-to-sequence design. The first model uses a Conformer-based encoder \cite{9414567} with a CTC/Attention decoder \cite{8068205}. The second model features a Transformer-based encoder (AV-HuBERT \cite{shi22_interspeech}) paired with a CTC/Attention decoder.

In the Conformer-based encoder (Figure \ref{fig:avsr_model_overview}, middle), the visual extraction module uses a modified ResNet-18, where the first layer is a spatiotemporal convolution with a kernel size of $5\times7\times7$ and a stride of $1\times2\times2$. This is followed by Conformer layers for deeper processing. Similarly, the audio extraction module consists of a 1D ResNet-18 followed by Conformer layers. The extracted visual and audio features are then concatenated and fused using a multi-layer perceptron (MLP).

In the AV-HuBERT encoder (Figure \ref{fig:avsr_model_overview}, right), the visual extraction module is similar to that of the Conformer-based encoder. However, the audio feature extraction is simplified to a linear feedforward network (FFN) projection layer. Unlike the Conformer-based encoder, where features are processed separately before fusion, AV-HuBERT concatenates and fuses audio and visual features before passing them through Transformer encoder layers.

The CTC/Attention decoder architecture \cite{8068205} consists of two parallel heads. The first head is a linear projection layer that directly predicts the target sequence from the encoder features and is trained using the CTC loss, which assumes conditional independence between each output prediction. The second head employs Transformer decoder layers followed by a linear layer to generate the target sequence, trained with cross-entropy (CE) loss. The final loss is a weighted sum of the CTC and CE losses, allowing the model to force monotonic alignments and at the same time get rid of the conditional independence assumption.



\vspace{-0.5em}
\section{Experimental setup}

\subsection{Data pre-processing}


We follow previous works \cite{shi22_interspeech, 9414567} for audio and video pre-processing. The training set consists of 269 hours of video from approximately 2475 speakers. The video is processed frame by frame, with each frame cropped to extract the mouth region of interest using a $96\times96$ bounding box. The audio is sampled at 16 kHz. For data augmentation, we apply horizontal flipping, random cropping, and adaptive time masking to the visual frames. For audio, we introduce background interference by adding up to two interfering speakers (random pick from the training set) with varying signal-to-noise ratios (SNR) set \(-5\), \(0\), \(5\), and \(10\) dB. Similar to \cite{9414567}, we further apply adaptive time masking to augment the audio data.


The original test set consists of approximately 2 hours of recordings from 36 speakers. While the evaluation data already includes naturally noisy samples with various noise types such as ``natural'', ``music'' and ``babble'', we aim to specifically assess the impact of the cocktail-party effect, where overlapping speech presents a greater challenge. In addition to the original test set (with zero interferer and SNR=$\infty$), we simulate cocktail-party conditions by introducing up to two interfering speakers, randomly selected from the test set, with SNR levels of \(-5\), \(0\), \(5\), and \(10\) dB, following the same approach used during training. The resulting ViCocktail test set comprises nine subsets with different (SNR, Interferer) conditions: ($\infty$,\(0\)), (\(-5\), \(1\)), (\(-5\), \(2\)), (\(0\), \(1\)), (\(0\), \(2\)), (\(5\), \(1\)), (\(5\), \(2\)), (\(10\), \(1\)), (\(10\), \(2\)). 

At SNR=$-5$dB, the target speaker, who is visible in the video, has significantly degraded audio, making their voice difficult to hear, while the interfering speaker is much clearer. In this case, distinguishing the target speaker based solely on auditory cues is nearly impossible. Conversely, at SNR=$10$dB, the target speaker's speech remains clear, though background speech from interfering speakers is still audible.

\vspace{-0.5em}
\subsection{Implementation details}

The Conformer CTC/Attention consists of 12 encoder layers for both audio and visual inputs, each with 16 attention heads. The CTC/Attention decoder is a 6-layer Transformer with the same dimensions and number of attention heads as the encoder. The AV-HuBERT CTC/Attention model uses the AV-HuBERT large as the encoder, which has 24 transformer blocks, each with 16 attention heads. The decoder CTC/Attention in this model is similar to the first, consisting of a 6-layer transformer. For the target vocabulary, we use SentencePiece \cite{kudo-2018-subword} subword units with a vocabulary size of 2057. Besides the audio-visual model architecture, we also experiment with audio-only and visual-only models to analyze the contribution of each modality. In the audio-only model, the visual extraction module and fusion layer are removed, while in the visual-only model, the audio extraction module and fusion layer are excluded.

In our experiments, we explore different parameter initializations for training our models. For the Conformer CTC/Attention model, we initialize parameters from Auto-AVSR \cite{10096889}, which was trained on 3,448 hours of English video data and achieved state-of-the-art performance on the LRS2 \cite{8099850} AVSR dataset. For the AV-HuBERT CTC/Attention model, we use pre-trained parameters from the MuAViC encoder model (which shares the same architecture as the AV-HuBERT encoder) \cite{anwar23_interspeech}, which has two versions: an English-only checkpoint trained on 436 hours of video data and a multilingual checkpoint trained on 708 hours of video data across eight languages (Arabic, German, Greek, Spanish, French, Italian, Portuguese, and Russian). Notably, the MuAViC encoder model was originally initialized from the AV-HuBERT large checkpoint, which was self-supervised on 1,759 hours of unlabeled video data.


\vspace{-1em}
\section{Results}

Table \ref{tab:experiment} presents the performance of different models on the ViCocktail dataset. Overall, multimodal audio-visual models (AV[1-4]) outperform both audio-only (A[1-5]) and visual-only models (V[1-5]), while audio-only models achieve better results than visual-only models.

Audio-only models are highly sensitive to noise, with WER increasing significantly as noise levels rise. The Wav2Vec2 ASR model (A1), used to generate automatic transcripts for the training dataset, performs well in clean conditions (SNR=$\infty$ with no interfering speakers), achieving a WER of 7.53\%. However, as noise levels increase and more interferers are introduced, A1's performance deteriorates drastically. In the most extreme case, with two interferers and SNR=$-5$, the WER rises 8.3 times, reaching 70.01\%. The audio-only model A2, a fine-tuned version of Whisper, exhibits a similar degradation pattern but performs worse than A1 overall. In contrast, the visual-only model (V1) is unaffected by noise, achieving its best WER of 41.34\%. While visual-only models underperform compared to audio-only models in clean conditions, they surpass almost all audio-only models in noisy environments.

Audio-visual models strike a balance between audio-only and visual-only models. The AV1 model achieves strong performance on the clean set, with a WER of 9.4\%, closely matching the audio-only A1 model's 7.53\% WER. Moreover, audio-visual models retain the robustness to noise observed in visual-only models, showing minimal degradation in noisy conditions. In the most extreme case, AV1's WER only doubles to 18.39\%, a significant improvement over the 8.3-fold increase (70.01\% WER) seen in the audio-only A1 model.

The AV-HuBERT CTC/Attention model outperforms the Conformer CTC/Attention model across all settings and modalities. This is noteworthy, as AV-HuBERT undergoes self-supervised training for cluster prediction before being fine-tuned for downstream tasks like AVSR, whereas the Conformer CTC/Attention model is directly trained for AVSR. Although both models are pre-trained on English, the self-supervised pre-training in AV-HuBERT may contribute to its greater robustness when adapting to Vietnamese.

When comparing different pre-training strategies for adapting to Vietnamese, the AV-HuBERT CTC/Attention model clearly benefits from pre-training, significantly outperforming training from scratch. Additionally, adaptation from the English checkpoint yields better results than the multilingual checkpoint. Similar observations hold for the Conformer CTC/Attention model, while the English pre-trained checkpoint provides reasonable performance, training this architecture from scratch fails to fit the data effectively.

\vspace{-0.5em}
\section{Conclusions}

In this work, we introduced a new AVSR dataset and established a strong baseline AVSR model for the Vietnamese language. Our data collection and preparation process is fully automated and can be extended to other languages. We benchmarked various architectures and pre-trained models to determine the most effective initialization, with AV-HuBERT providing the best performance for Vietnamese AVSR. In the future, we plan to scale up data collection and explore self-supervised pre-training to develop more efficient models. The data pipeline, AVSR dataset, and baseline model are made publicly available for further research at: https://github.com/nguyenvulebinh/viCocktail


\bibliographystyle{IEEEtran}
\bibliography{mybib}

\begin{thebibliography}{10}
\providecommand{\url}[1]{#1}
\csname url@samestyle\endcsname
\providecommand{\newblock}{\relax}
\providecommand{\bibinfo}[2]{#2}
\providecommand{\BIBentrySTDinterwordspacing}{\spaceskip=0pt\relax}
\providecommand{\BIBentryALTinterwordstretchfactor}{4}
\providecommand{\BIBentryALTinterwordspacing}{\spaceskip=\fontdimen2\font plus
\BIBentryALTinterwordstretchfactor\fontdimen3\font minus \fontdimen4\font\relax}
\providecommand{\BIBforeignlanguage}[2]{{%
\expandafter\ifx\csname l@#1\endcsname\relax
\typeout{** WARNING: IEEEtran.bst: No hyphenation pattern has been}%
\typeout{** loaded for the language `#1'. Using the pattern for}%
\typeout{** the default language instead.}%
\else
\language=\csname l@#1\endcsname
\fi
#2}}
\providecommand{\BIBdecl}{\relax}
\BIBdecl

\bibitem{mcgurk1976hearing}
H.~McGurk and J.~MacDonald, ``Hearing lips and seeing voices,'' \emph{Nature}, vol. 264, no. 5588, pp. 746--748, 1976.

\bibitem{41402}
B.~Yuhas, M.~Goldstein, and T.~Sejnowski, ``Integration of acoustic and visual speech signals using neural networks,'' \emph{IEEE Communications Magazine}, vol.~27, no.~11, pp. 65--71, 1989.

\bibitem{duchnowski94_icslp}
P.~Duchnowski, U.~Meier, and A.~Waibel, ``See me, hear me: integrating automatic speech recognition and lip-reading,'' in \emph{3rd International Conference on Spoken Language Processing (ICSLP 1994)}, 1994, pp. 547--550.

\bibitem{waibel1996multimodal}
A.~Waibel, M.~T. Vo, P.~Duchnowski, and S.~Manke, ``Multimodal interfaces,'' \emph{Artificial Intelligence Review}, vol.~10, pp. 299--319, 1996.

\bibitem{8099850}
J.~S. Chung, A.~Senior, O.~Vinyals, and A.~Zisserman, ``Lip reading sentences in the wild,'' in \emph{2017 IEEE Conference on Computer Vision and Pattern Recognition (CVPR)}, 2017, pp. 3444--3453.

\bibitem{afouras2018lrs3tedlargescaledatasetvisual}
\BIBentryALTinterwordspacing
T.~Afouras, J.~S. Chung, and A.~Zisserman, ``Lrs3-ted: a large-scale dataset for visual speech recognition,'' 2018. [Online]. Available: \url{https://arxiv.org/abs/1809.00496}
\BIBentrySTDinterwordspacing

\bibitem{shillingford19_interspeech}
B.~Shillingford, Y.~Assael, M.~W. Hoffman, T.~Paine, C.~Hughes, U.~Prabhu, H.~Liao, H.~Sak, K.~Rao, L.~Bennett, M.~Mulville, M.~Denil, B.~Coppin, B.~Laurie, A.~Senior, and N.~de~Freitas, ``Large-scale visual speech recognition,'' in \emph{Interspeech 2019}, 2019, pp. 4135--4139.

\bibitem{6920467}
S.~Antar, A.~Sagheer, S.~Aly, and M.~F. Tolba, ``Avas: Speech database for multimodal recognition applications,'' in \emph{13th International Conference on Hybrid Intelligent Systems (HIS 2013)}, 2013, pp. 123--128.

\bibitem{ELREFAEI2019400}
\BIBentryALTinterwordspacing
L.~A. Elrefaei, T.~Q. Alhassan, and S.~S. Omar, ``An arabic visual dataset for visual speech recognition,'' \emph{Procedia Computer Science}, vol. 163, pp. 400--409, 2019, 16th Learning and Technology Conference 2019Artificial Intelligence and Machine Learning: Embedding the Intelligence. [Online]. Available: \url{https://www.sciencedirect.com/science/article/pii/S1877050919321611}
\BIBentrySTDinterwordspacing

\bibitem{10095796}
C.~Chen, D.~Wang, and T.~F. Zheng, ``Cn-cvs: A mandarin audio-visual dataset for large vocabulary continuous visual to speech synthesis,'' in \emph{ICASSP 2023 - 2023 IEEE International Conference on Acoustics, Speech and Signal Processing (ICASSP)}, 2023, pp. 1--5.

\bibitem{Chen2020LipreadingWD}
\BIBentryALTinterwordspacing
X.~Chen, J.~Du, and H.~Zhang, ``Lipreading with densenet and resbi-lstm,'' \emph{Signal, Image and Video Processing}, vol.~14, pp. 981 -- 989, 2020. [Online]. Available: \url{https://api.semanticscholar.org/CorpusID:214376123}
\BIBentrySTDinterwordspacing

\bibitem{10.1145/3338533.3366579}
\BIBentryALTinterwordspacing
Y.~Zhao, R.~Xu, and M.~Song, ``A cascade sequence-to-sequence model for chinese mandarin lip reading,'' in \emph{Proceedings of the 1st ACM International Conference on Multimedia in Asia}, ser. MMAsia '19.\hskip 1em plus 0.5em minus 0.4em\relax New York, NY, USA: Association for Computing Machinery, 2020. [Online]. Available: \url{https://doi.org/10.1145/3338533.3366579}
\BIBentrySTDinterwordspacing

\bibitem{ivanko-etal-2022-rusavic}
\BIBentryALTinterwordspacing
D.~Ivanko, A.~Axyonov, D.~Ryumin, A.~Kashevnik, and A.~Karpov, ``{RUSAVIC} corpus: {R}ussian audio-visual speech in cars,'' in \emph{Proceedings of the Thirteenth Language Resources and Evaluation Conference}.\hskip 1em plus 0.5em minus 0.4em\relax Marseille, France: European Language Resources Association, Jun. 2022, pp. 1555--1559. [Online]. Available: \url{https://aclanthology.org/2022.lrec-1.166/}
\BIBentrySTDinterwordspacing

\bibitem{Verkhodanova2016HAVRUSCH}
\BIBentryALTinterwordspacing
V.~Verkhodanova, A.~L. Ronzhin, I.~S. Kipyatkova, D.~Ivanko, A.~Karpov, and M.~Železn{\'y}, ``Havrus corpus: High-speed recordings of audio-visual russian speech,'' in \emph{International Conference on Speech and Computer}, 2016. [Online]. Available: \url{https://api.semanticscholar.org/CorpusID:34549549}
\BIBentrySTDinterwordspacing

\bibitem{7961743}
A.~Fernandez-Lopez, O.~Martinez, and F.~M. Sukno, ``Towards estimating the upper bound of visual-speech recognition: The visual lip-reading feasibility database,'' in \emph{2017 12th IEEE International Conference on Automatic Face \& Gesture Recognition (FG 2017)}, 2017, pp. 208--215.

\bibitem{data8010015}
\BIBentryALTinterwordspacing
A.~Berkol, T.~Tümer-Sivri, N.~Pervan-Akman, M.~Çolak, and H.~Erdem, ``Visual lip reading dataset in turkish,'' \emph{Data}, vol.~8, no.~1, 2023. [Online]. Available: \url{https://www.mdpi.com/2306-5729/8/1/15}
\BIBentrySTDinterwordspacing

\bibitem{10.1007/978-3-319-54184-6_6}
J.~S. Chung and A.~Zisserman, ``Lip reading in the wild,'' in \emph{Computer Vision -- ACCV 2016}, S.-H. Lai, V.~Lepetit, K.~Nishino, and Y.~Sato, Eds.\hskip 1em plus 0.5em minus 0.4em\relax Cham: Springer International Publishing, 2017, pp. 87--103.

\bibitem{shi22_interspeech}
B.~Shi, W.-N. Hsu, and A.~Mohamed, ``Robust self-supervised audio-visual speech recognition,'' in \emph{Interspeech 2022}, 2022, pp. 2118--2122.

\bibitem{rouditchenko24_interspeech}
A.~Rouditchenko, Y.~Gong, S.~Thomas, L.~Karlinsky, H.~Kuehne, R.~Feris, and J.~Glass, ``Whisper-flamingo: Integrating visual features into whisper for audio-visual speech recognition and translation,'' in \emph{Interspeech 2024}, 2024, pp. 2420--2424.

\bibitem{10096889}
P.~Ma, A.~Haliassos, A.~Fernandez-Lopez, H.~Chen, S.~Petridis, and M.~Pantic, ``Auto-avsr: Audio-visual speech recognition with automatic labels,'' in \emph{ICASSP 2023 - 2023 IEEE International Conference on Acoustics, Speech and Signal Processing (ICASSP)}, 2023, pp. 1--5.

\bibitem{babu22_interspeech}
A.~Babu, C.~Wang, A.~Tjandra, K.~Lakhotia, Q.~Xu, N.~Goyal, K.~Singh, P.~{von Platen}, Y.~Saraf, J.~Pino, A.~Baevski, A.~Conneau, and M.~Auli, ``Xls-r: Self-supervised cross-lingual speech representation learning at scale,'' in \emph{Interspeech 2022}, 2022, pp. 2278--2282.

\bibitem{xu22b_interspeech}
Q.~Xu, A.~Baevski, and M.~Auli, ``Simple and effective zero-shot cross-lingual phoneme recognition,'' in \emph{Interspeech 2022}, 2022, pp. 2113--2117.

\bibitem{zhang2017s3fd}
S.~Zhang, X.~Zhu, Z.~Lei, H.~Shi, X.~Wang, and S.~Z. Li, ``S3fd: Single shot scale-invariant face detector,'' in \emph{Proceedings of the IEEE international conference on computer vision}, 2017, pp. 192--201.

\bibitem{Liao_2023_CVPR}
J.~Liao, H.~Duan, K.~Feng, W.~Zhao, Y.~Yang, and L.~Chen, ``A light weight model for active speaker detection,'' in \emph{Proceedings of the IEEE/CVF Conference on Computer Vision and Pattern Recognition (CVPR)}, June 2023, pp. 22\,932--22\,941.

\bibitem{Chung2016OutOT}
\BIBentryALTinterwordspacing
J.~S. Chung and A.~Zisserman, ``Out of time: Automated lip sync in the wild,'' in \emph{ACCV Workshops}, 2016. [Online]. Available: \url{https://api.semanticscholar.org/CorpusID:26294509}
\BIBentrySTDinterwordspacing

\bibitem{king2009dlib}
D.~E. King, ``Dlib-ml: A machine learning toolkit,'' \emph{The Journal of Machine Learning Research}, vol.~10, pp. 1755--1758, 2009.

\bibitem{Thai_Binh_Nguyen_wav2vec2_vi_2021}
\BIBentryALTinterwordspacing
T.~B. Nguyen, ``{Vietnamese end-to-end speech recognition using wav2vec 2.0},'' 09 2021. [Online]. Available: \url{https://github.com/vietai/ASR}
\BIBentrySTDinterwordspacing

\bibitem{10446589}
T.-B. Nguyen and A.~Waibel, ``Synthetic conversations improve multi-talker asr,'' in \emph{ICASSP 2024 - 2024 IEEE International Conference on Acoustics, Speech and Signal Processing (ICASSP)}, 2024, pp. 10\,461--10\,465.

\bibitem{10889116}
T.-B. Nguyen and A.Waibel, ``Msa-asr: Efficient multilingual speaker attribution with frozen asr models,'' in \emph{ICASSP 2025 - 2025 IEEE International Conference on Acoustics, Speech and Signal Processing (ICASSP)}, 2025, pp. 1--5.

\bibitem{le2024phowhisper}
\BIBentryALTinterwordspacing
T.-T. Le, L.~T. Nguyen, and D.~Q. Nguyen, ``Phowhisper: Automatic speech recognition for vietnamese,'' in \emph{The Second Tiny Papers Track at ICLR 2024}, 2024. [Online]. Available: \url{https://openreview.net/forum?id=x3c3MkJfpG}
\BIBentrySTDinterwordspacing

\bibitem{9414567}
P.~Ma, S.~Petridis, and M.~Pantic, ``End-to-end audio-visual speech recognition with conformers,'' in \emph{ICASSP 2021 - 2021 IEEE International Conference on Acoustics, Speech and Signal Processing (ICASSP)}, 2021, pp. 7613--7617.

\bibitem{8068205}
S.~Watanabe, T.~Hori, S.~Kim, J.~R. Hershey, and T.~Hayashi, ``Hybrid ctc/attention architecture for end-to-end speech recognition,'' \emph{IEEE Journal of Selected Topics in Signal Processing}, vol.~11, no.~8, pp. 1240--1253, 2017.

\bibitem{kudo-2018-subword}
\BIBentryALTinterwordspacing
T.~Kudo, ``Subword regularization: Improving neural network translation models with multiple subword candidates,'' in \emph{ACL}.\hskip 1em plus 0.5em minus 0.4em\relax Melbourne, Australia: Association for Computational Linguistics, Jul. 2018, pp. 66--75. [Online]. Available: \url{https://aclanthology.org/P18-1007/}
\BIBentrySTDinterwordspacing

\bibitem{anwar23_interspeech}
M.~Anwar, B.~Shi, V.~Goswami, W.-N. Hsu, J.~Pino, and C.~Wang, ``Muavic: A multilingual audio-visual corpus for robust speech recognition and robust speech-to-text translation,'' in \emph{Interspeech 2023}, 2023, pp. 4064--4068.

\end{thebibliography}

\end{document}